%% file: main.tex
\documentclass[runningheads]{llncs}

\usepackage{eccv}

\usepackage{eccvabbrv}

\usepackage{microtype}
\usepackage{graphicx}
\usepackage{subcaption}
\usepackage{booktabs} 
\usepackage{pgf-pie} 
\usepackage{xcolor}
\usepackage{tikz}
\usepackage{multirow}
\usepackage{multicol}
\usepackage{pgfplots}
\usepackage{caption}
\pgfplotsset{compat=1.18}
\usepackage{makecell}
\usepackage{threeparttable}
\usepackage{float}    
\usepackage{array}    
\usepackage[utf8]{inputenc}
\usepackage[T1]{fontenc}
\usepackage{mdframed}
\usepackage{tabularx}  
\usepackage{enumitem}  
\usepackage{placeins} 

\makeatletter
\renewcommand\subsection{\@startsection{subsection}{2}{\z@}%
  {-8\p@ \@plus -2\p@ \@minus -2\p@}%
  {4\p@ \@plus 2\p@ \@minus 2\p@}%
  {\normalfont\large\bfseries\boldmath}}
\renewcommand\subsubsection{\@startsection{subsubsection}{3}{\z@}%
  {-6\p@ \@plus -2\p@ \@minus -2\p@}%
  {0.5em \@plus 0.22em \@minus 0.1em}%
  {\normalfont\normalsize\bfseries\boldmath}}
\makeatother

\usepackage{hyperref}

\usepackage{orcidlink}

\usepackage{amsmath}
\usepackage{amssymb}
\usepackage{mathtools}

\usepackage[capitalize,noabbrev]{cleveref}

\usepackage[textsize=tiny]{todonotes}

\begin{document}

\title{LVSum: A Benchmark for Timestamp-Aware Long Video Summarization}

\titlerunning{LVSum: A Benchmark for Timestamp-Aware Long Video Summarization}

\author{Alkesh Patel\thanks{Equal contribution.} \and
Melis Ozyildirim\protect\footnotemark[1] \and
 Ying-Chang Cheng \and Ganesh Nagarajan}

\authorrunning{A. Patel et al.}

\institute{Apple\\
\email{\{alkesh.patel, melisozyildirim, ycheng27, ganesh.nagarajan\}@apple.com}}



\maketitle

\begin{abstract}
Long video summarization presents significant challenges for multimodal large language models (MLLMs), particularly in maintaining temporal fidelity over extended durations and producing summaries that are both semantically and temporally grounded. We introduce LVSum, a human-annotated benchmark for evaluating \textbf{long-form} video summarization with fine-grained temporal alignment. LVSum comprises \textbf{72 diverse videos} spanning \textbf{13 domains} with an average duration of 16 minutes, each annotated with \textbf{up to 10 human-generated summaries} containing temporal references. We conduct a comprehensive evaluation of leading proprietary and open-source MLLMs using newly introduced LLM-based metrics for \textbf{content relevance} and \textbf{modality coherence}, alongside standard automatic metrics. Our experiments reveal three key findings: (1) transcripts contribute substantially more
to summarization quality than visual frames alone, (2) a significant performance gap persists between model-generated and human-written summaries, and (3) current MLLMs exhibit systematic weaknesses in temporal grounding, instruction adherence, and cross-modal coherence. We release the dataset and code.

\end{abstract}

\section{Introduction}

Video contents are increasingly created and consumed by millions of users, creating a need to effectively make them accessible at the right time and in the right form to reduce cognitive load and enable fast decision making. Long-form videos -- such as lectures, documentaries, news reports, vlogs, and podcasts -- have become ubiquitous across platforms.  However, efficiently summarizing their content remains a challenge for both humans and machines. Long video summarization aims to condense extended video content into concise, coherent representations, often including temporally grounded summaries that highlight what happened and when. While recent advances in multimodal large language models (MLLMs) have shown promise in video understanding, temporal precision and scalability over long video sequences remain key limitations.

Despite the rapid growth of multimodal large language models (MLLMs), existing methods fall short in several major ways:

\begin{enumerate}
    \item \textbf{Lack of temporal grounding.}
    Most current MLLMs can summarize what happened in a video, but struggle to predict \emph{when} important moments occurred. They lack explicit temporal supervision, such as timestamped key moments or event ranges.

    \item \textbf{Limited context window for long videos.}
    Long videos (ranging from tens of minutes to several hours) pose significant scaling challenges due to context length limitations. As a result, MLLMs often truncate inputs or rely on sparse frame sampling, leading to loss of temporal continuity and context.

    \item \textbf{Weak instruction following for summarization tasks.}
    Many MLLMs are trained on generic instruction-tuned corpora but are not fine-tuned for temporal video summarization. The lack of domain- or task-specific instruction tuning often results in less targeted outputs. In particular, constraints such as summary length budgets and category-dependent guidance play a critical role in producing high-quality summaries.

    \item \textbf{Lack of holistic metrics.}
    Existing automatic metrics (e.g., F1, Kendall's $\tau$, Spearman's $\rho$) inadequately capture semantic relevance, temporal coherence, and category-dependent preferences in video summaries. Human evaluation further introduces subjectivity and scalability challenges, highlighting the need for more comprehensive and reliable evaluation protocols.
\end{enumerate}

To address some of these challenges, we introduce LVSum, a benchmark focused on long-duration video summarization, featuring 72 videos with an average duration of 16 minutes and human-annotated summaries with fine-grained temporal alignment. LVSum spans over 13 categories (\textit{e.g., Lectures/Tutorials, Documentaries, News Reports, TV Shows, Podcasts, Vlogs, Webinars, Comedy, Explainer Videos, Games, Memes, Reactions, Debates/Panel Discussions}), and is designed to rigorously evaluate a model's ability to produce coherent, timestamp-anchored summaries from lengthy video inputs.

Finally, we conduct a systematic evaluation of proprietary and open-source MLLMs — including models like 
Opus-4.5 ~\cite{anthropic2025opus45}, Gemini-2.5-Pro ~\cite{gemini25report2025}, and open-source variants like Qwen3-VL-235B-A22B-Instruct ~\cite{qwen3vlreport2025}-- on LVSum. We assess them across multiple dimensions: semantic relevance, temporal accuracy, modality alignment, and scalability to long input sequences. Our results uncover critical gaps in current models' abilities to track narrative flow over time.


This paper makes three primary contributions:
\begin{itemize}
    
    
    \item \textbf{LVSum}\footnote{The dataset and corresponding source code will be made publicly available upon acceptance of the paper.}, a new benchmark explicitly designed for \textbf{long-duration} video summarization, featuring \textbf{72 videos} with up to \textbf{10 human-annotated summaries} per video and fine-grained temporal alignment across diverse domains.
    
    \item \textbf{Two new evaluation metrics — Modality Coherence (MC) and Content Relevance (CR)} — designed to quantify cross-modal grounding and semantic faithfulness in long-video summarization, addressing limitations of existing rank-based correlation metrics.
    
    \item \textbf{A large-scale empirical study of proprietary and open-source MLLMs} on long video summarization, combining quantitative benchmarking with systematic failure-mode analysis to uncover limitations in temporal grounding, instruction adherence, and cross-modal reasoning under extended contexts.

\end{itemize}
We hope LVSum and our findings will serve as a foundation for future research in temporally-aware multimodal learning and evaluation.

\begin{table*}[t]
  \caption{Comparison of LVSum with other Video Summarization datasets}
  \label{tab:lvsum-benchmark-comparison}
  \begin{center}
  \begin{threeparttable}
    \setlength{\tabcolsep}{2.5pt}
    \begin{footnotesize}
      \begin{sc}
      \setlength{\tabcolsep}{5pt}
        \begin{tabular}{lccccc}
          \toprule
          Dataset  & Domain & \# Videos  &  \makecell{Len.\\(avg)}     & \makecell{Ann./\\video} & \makecell{Imp.\\Score}\\ 
          \midrule
          SumMe & 3 cat. & 25 & 1--3 (2.6)m & 15--18 & $\checkmark$ \\ 
          TVSum & 10 cat. & 50 & 2--10 (4.2)m & 20 & $\checkmark$ \\
          OVP & 5 cat. & 50 & 1--4 (3)m & 5 & $\times$ \\
          YouTube & 5 cat. & 50 & 1--10 (3.5)m & 5 & $\times$ \\ 
          EDUVSUM & Lectures & 98 & up to 15 (6.5)m & 1 & $\checkmark$ \\ 
          VideoXum & open & 14K & up to 12.5 (2)m & 10 & $\times$ \\
          Instruct-V2Xum & open & 30K & 1-15 (3)m & 1 &  $\times$ \\
          LVSum & 13 cat. & 72 & \textbf{10--55 (16)m} & 10 & $\checkmark$ \\ 
          \bottomrule
        \end{tabular}
      \end{sc}
    \end{footnotesize}
  \end{threeparttable}
  \end{center}
\end{table*}

\section{Related Work}
\textbf{Video summarization tasks}\\
There are several task variants for video summarization. Video-to-video (V2V) summarization ~\cite{deAvila2011VSUMM, gygli2014creating, Song2015TVSumSW,ghauri2020eduvsum} condenses a long video into a shorter version preserving its most informative and diverse content.  This is different from Video-to-text (V2T) summarization ~\cite{chen2017video, mylonas2025sdvsum, liu-etal-2025-talk}, which generates a compact textual summary for a video. Query-driven video (Q2V) summarization  ~\cite{Sharghi2016QueryFocused, wei2018video, huang2020query, Narasimhan2021CLIPIt, narasimhan2022tl, he2023align} produces user-customized summaries. Video-to-video-text (V2VT) summarization ~\cite{lin2023videoxum, hua2025v2xum} provides both video and text summaries.  In this paper, we focus on the V2VT task, where the input is a video with its transcript and the output consists of timestamped video intervals with corresponding descriptions. Unless otherwise specified, we use ``video summarization'' to refer to V2VT.\\
\textbf{Video Summarization Datasets}\\
SumMe ~\cite{gygli2014creating} and TVSum ~\cite{Song2015TVSumSW} have remained classic datasets for over a decade, but they are dominated by shorter web/user videos compared to modern long-form videos across many categories. Moreover, their volume remains small (25-50 videos). More recent datasets such as VideoXum ~\cite{lin2023videoxum}  provides large scale dataset repurposing the ActivityNetCap ~\cite{krishna2017dense}. Instruct-V2Xum ~\cite{hua2025v2xum} is another large video summarization dataset created using large language model-based summaries; however, it lacks high-quality human annotations. Additionally, its summaries are represented as selected keyframes rather than contiguous temporal intervals. All prior datasets contain short videos ($\leq$ 10 mins). In contrast, we focus on long videos with high-quality human-annotated key-shots paired with short descriptions to better reflect real-world, summary-worthy content.\\
\textbf{Video Summarization with LLMs}\\
Video summarization has transitioned from extractive methods, which select keyframes based on visual saliency or diversity, to abstractive and semantic-aware approaches \cite{Apostolidis2021VideoSU}. Recent progress leverages MLLMs for understanding and describing video content. LLMVS ~\cite{lee2025video} adopts an LLM-based frame selection strategy to generate human-aligned summaries, while V2Xum-LLM ~\cite{hua2025v2xum} introduces temporal prompt instruction tuning for cross-modal summarization across video-to-video, video-to-text, and mixed settings. While promising, these methods are primarily evaluated on short videos, leaving the capability of MLLMs for long-video summarization largely untested.
To address this gap, we introduce LVSum, a long-video summarization dataset and benchmark designed to evaluate state-of-the-art MLLMs under realistic constraints. LVSum includes long videos with rich human annotations, summary budgets, and evaluation protocols, enabling systematic benchmarking of modern MLLMs for long-video summarization.\\
\textbf{Evaluation protocols and metrics}\\
Reliable evaluation of video summarization depends on annotation richness and appropriate metrics. Early benchmarks such as SumMe and TVSum provided 15–20 human annotations per video, enabling researchers to model perceptual variance, but their limited scale (25–50 videos) is inadequate for evaluating modern multimodal LLMs. Beyond annotation density, evaluation metrics have also evolved. ~\cite{otani2019rethinking} showed that the widely used F1 score is strongly biased by segment-length distributions and argued for rank-based measures such as Kendall's $\tau$ and Spearman's $\rho$ to better capture importance. For cross-modal settings, VideoXum ~\cite{lin2023videoxum} introduced VT-CLIPScore to measure semantic consistency between video and text, and V2Xum-LLM ~\cite{hua2025v2xum} extended this direction with CLIP-based cross-modal metrics.
However, these recent metrics primarily evaluate semantic relevance and modal similarity, and overlook a new challenge introduced by MLLM-based summarization: modality coherence between predicted timestamps and their generated descriptions. In practice, we observe that MLLMs often produce plausible textual descriptions that do not correspond to the visual content of the predicted interval, creating cross-modal hallucinations that current metrics fail to capture. To address this gap, we incorporate MLLM-as-judge metrics that assess content relevance and introduce a modality coherence metric to penalize timestamp–description mismatch in V2VT summarization.

\section{Dataset for Long Video Summarization: LVSum}

\begin{figure}[t]
\centering
\definecolor{c1}{HTML}{4E79A7}
\definecolor{c2}{HTML}{F28E2B}
\definecolor{c3}{HTML}{E15759}
\definecolor{c4}{HTML}{76B7B2}
\definecolor{c5}{HTML}{59A14F}
\definecolor{c6}{HTML}{EDC949}
\definecolor{c7}{HTML}{AF7AA1}
\definecolor{c8}{HTML}{FF9DA7}
\definecolor{c9}{HTML}{9C755F}
\definecolor{c10}{HTML}{BAB0AC}
\definecolor{c11}{HTML}{8CD17D}
\definecolor{c12}{HTML}{B6992D}
\definecolor{c13}{HTML}{D37295}

\begin{minipage}{0.55\linewidth}
\centering
\begin{tikzpicture}
  \pie[
    radius=2.2,
    sum=auto,
    hide number,
    text=,
    color={c1,c2,c3,c4,c5,c6,c7,c8,c9,c10,c11,c12,c13}
  ]
  {
      28/Lectures,
      16/Podcasts,
      5/Vlogs,
      4/News Reports,
      4/Explainer Videos,
      3/Documentaries,
      3/Comedy,
      2/Animation,
      2/Debates,
      2/Reactions,
      1/TV Shows,
      1/Games,
      1/Memes
  }
\end{tikzpicture}
\end{minipage}
\hfill
\begin{minipage}{0.4\linewidth}
\small
\renewcommand{\arraystretch}{0.9}

\begin{tabular}{ll}
\colorbox{c1}{\phantom{x}} & Lectures \\
\colorbox{c2}{\phantom{x}} & Podcasts \\
\colorbox{c3}{\phantom{x}} & Vlogs \\
\colorbox{c4}{\phantom{x}} & News Reports \\
\colorbox{c5}{\phantom{x}} & Explainer Videos \\
\colorbox{c6}{\phantom{x}} & Documentaries \\
\colorbox{c7}{\phantom{x}} & Comedy \\
\colorbox{c8}{\phantom{x}} & Animation \\
\colorbox{c9}{\phantom{x}} & Debates \\
\colorbox{c10}{\phantom{x}} & Reactions \\
\colorbox{c11}{\phantom{x}} & TV Shows \\
\colorbox{c12}{\phantom{x}} & Games \\
\colorbox{c13}{\phantom{x}} & Memes \\
\end{tabular}

\end{minipage}

\caption{Distribution of video categories in the LVSum dataset.}
\label{fig:data_statistics}
\end{figure}

\subsection{Dataset Construction}


We begin by crawling approximately 4{,}000 videos from the web, each with a duration of at least 10 minutes. To ensure content diversity, we use an MLLM (Gemini-2.5-Pro) to assign open-ended semantic category labels to each video. These labels are then clustered, also using Gemini-2.5-Pro, into 13 higher-level categories to form a coarse taxonomy of video content. From this pool, we select 100 videos using weighted sampling proportional to the observed category distribution. This strategy preserves the natural long-tailed distribution of real-world video content while avoiding over-representation of dominant categories that would result from uniform sampling. The selected videos are then sent to human annotators for summarization. Annotators are required to be fluent in English, as the videos are in English, and all annotations are conducted by annotators based in the United States, Europe, and India to ensure linguistic proficiency and cultural diversity.

Annotators were instructed to filter out videos containing personally identifiable or sensitive content before proceeding. For valid videos, summarization was performed using the following protocol (See Appendix~\ref{appendix:annotation_guidelines} for more details): (1) watch the entire video to understand its content and narrative structure; (2) rewatch and identify crucial segments; (3) mark each segment with start/end timestamps, assign a concise textual description, and rate its relevance on a 1–3 scale; (4) iteratively refine segments while keeping the total interval length within 15\% of the video duration, a budget commonly adopted in prior video summarization benchmarks to balance coverage and conciseness; and (5) submit the final annotations.
Each video was annotated by 10 independent annotators. We manually reviewed submissions and discarded annotations that violated the instructions or the length budget.

Of the initial 100 videos, 11 were filtered due to personally identifiable or sensitive content, and 17 were removed due to updated content usage restrictions from source websites, leaving \textbf{72 videos} that constitute the \textbf{LVSum dataset}. Figure~\ref{fig:data_statistics} shows the category distribution across the 13 higher-level categories.

\subsection{Comparison with other video summarization dataset}
Table~\ref{tab:lvsum-benchmark-comparison} compares LVSum with widely used video summarization datasets. LVSum targets long-form summarization and contains 72 videos ranging from \textbf{10–55 minutes (avg. 16 min)}, substantially longer than any prior dataset. Each video includes up to 10 independent human annotations with interval-level importance scores, providing dense perceptual supervision and enabling reliable measurement of human agreement--an essential property for evaluating subjective summarization quality. No existing dataset jointly offers, (1) long videos, (2) multiple human references, and (3) explicit interval-level importance scores at scale.  VideoXum~\cite{lin2023videoxum} contains more videos but inherits structural limitations from ActivityNet Captions, including redundant dense captions and the absence of explicit importance scores, making it an incomplete proxy for real-world summarization. Instruct-V2Xum~\cite{hua2025v2xum} provides large-scale frame-level annotations but only a single reference per video, and lacks contiguous key-shot intervals that can be stitched into coherent summaries. EDUVSUM~\cite{ghauri2020eduvsum} similarly offers only one expert annotation and is domain-restricted to lecture videos, limiting both robustness and generalization. LVSum fills a critical gap by providing long-form videos, multi-reference annotations, and interval-level importance supervision, enabling meaningful benchmarking of modern MLLMs for subjective video summarization.

\section{Experiments and Results}
\label{experiments_results}

\subsection{Models Used}
To benchmark LVSum against state-of-the-art multimodal LLMs, we evaluate three representative systems: 
Sonnet-4.5, Gemini-2.5-Pro, and Qwen3-VL-235B-A22B-Instruct (Qwen3-VL-235B) --reflecting both proprietary and open-source systems. Since nearly all videos include spoken narration or dialogue, we generate timestamped transcripts using 
Gemini-2.5-Pro. Each model receives 96 uniformly sampled frames along with the transcript as input. The prompt template used for summary prediction is shown in \hyperlink{appendix:prompt_video_summarization}{Appendix~\ref*{appendix:prompt_video_summarization}}.

\subsection{Evaluation Metrics}
In practice, current MLLMs are prone to producing summary segments whose accumulated duration exceeds the target 15\% budget, even when explicitly instructed with a length constraint. Moreover,~\cite{otani2019rethinking} showed that Kendall's $\tau$ and Spearman's $\rho$ are more robust than F1-based metrics when segment lengths vary. For these reasons, we use Kendall's $\tau$ and Spearman's $\rho$ as our primary metrics and report the average predicted summary length to contextualize metric values.

Because LVSum contains long videos, we compute importance rankings at \textbf{second-level granularity} rather than frame-level granularity used in short-video benchmarks such as SumMe and TVSum. We uniformly subsample videos at 1 fps and assign segment-level saliency scores at the corresponding timestamps. This resolution aligns with the temporal precision of MLLMs, which typically generate timestamps at the level of seconds rather than frames, and provides a more faithful comparison for long-video summarization.

We also introduce MLLM-as-Judge metrics for content relevance and modality coherence:

\textbf{Content Relevance (CR)}: This metric measures how well the generated summary captures the important information from the video relative to a ground-truth text summary corresponding to annotated key moments. The main intuition is to verify: \textit{Did the model select or describe the right events? Does it omit irrelevant or trivial segments? Are the semantic concepts, objects, interactions, and outcomes aligned with the reference annotations?}
Each summary is rated on a scale from 1 (poor relevance) to 5 (high relevance), with higher scores indicating stronger semantic alignment with the ground truth.

\textbf{Modality Coherence (MC)}: This metric evaluates whether the predicted summary is cross-modally consistent within the predicted summary intervals and does not hallucinate content unsupported by the corresponding video segments. Given a predicted summary consisting of timestamped segments with descriptions, we extract the visual and auditory content from each predicted interval and compare them with the associated textual description. MC therefore asks whether the visual, auditory, and textual cues for a segment agree: \textit{Does the segment show what the description claims? Is speech or sound described that does not exist in video? Are actions or objects hallucinated?}
Each segment is scored on a scale from 1 (severe cross-modal inconsistency or hallucination) to 5 (strong cross-modal consistency), and scores are averaged across segments to obtain the final MC score.

We use Gemini-2.5-Pro ~\cite{OpenAI2025GPT5Science} as judge. The prompts used for MLLM-as-Judge metrics are provided in \hyperlink{appendix:mllm-as-judge}{Appendix ~\ref*{appendix:mllm-as-judge}}.

\subsection{Baselines}
We provide following simple baselines:\\
\textbf{1. Random interval baseline:} We randomly sample a set of temporal segments for each video, where segment lengths are drawn from \{3, 4, 5\} seconds. This range is chosen to reflect the typical perceptual duration of a visually or semantically meaningful moment in long-form videos. Each segment is assigned a random importance score from \{1,2,3\}. The total duration of sampled segments is constrained to $\approx15\%$ of the video length to match the summary budget. \\
\textbf{2. Evenly spaced interval baseline:} We uniformly space temporal segments of \{3, 4, 5\}-seconds lengths across each video and assign importance scores uniformly at random from \{1, 2, 3\}. The total duration is again constrained to $\approx15\%$ of the video length.

These baselines establish lower bounds for temporal selection and highlight the contribution of visual and multimodal cues in MLLM-based summarization.

\section{Results and Discussion}
\subsection{Main Results}
Table~\ref{tab:lvsum-benchmark} summarizes model performance on LVSum. Using a leave-one-out protocol across annotators, human agreement reaches $\tau=0.13388$ and $\rho=0.13635$, providing an empirical upper bound. Among models, Opus-4.5 achieves the highest temporal agreement ($\tau=0.10093$, $\rho=0.10513$), reaching approximately 75\% of human-level performance. Gemini-2.5-Pro follows closely in rank correlation ($\tau=0.08902$, $\rho=0.09071$), while Qwen3-VL-235B lags behind. However, ranking accuracy alone does not reflect true summarization behavior. Both Opus-4.5 (30\%) and Qwen3-VL-235B (56\%) substantially exceed the 15\% summary budget, indicating a tendency toward over-coverage rather than temporal compression. In contrast, Gemini-2.5-Pro adheres closely to the budget (12\%) while maintaining competitive rank alignment. All MLLMs achieve high Content Relevance (CR), suggesting strong semantic understanding. However, Gemini-2.5-Pro attains the highest Modality Coherence (MC = 4.32), indicating better grounding between predicted timestamps and corresponding video/audio content. Overall, these results highlight three emerging failure modes for MLLM-based long-video summarization: \textbf{(i) over-coverage}, where models inflate summary length to increase content retention; \textbf{(ii) temporal misalignment}, where semantic importance is captured but timestamps are imprecise; and \textbf{(iii) cross-modal mismatch}, where descriptions hallucinate actions or objects not supported by the corresponding video segment. Together, these findings suggest that while MLLMs are capable of identifying semantically important events, they do not yet exhibit robust summarization behavior that balances content relevance, temporal compression, and multimodal grounding.

\begin{table*}[t]
  \caption{Evaluation of representative models on the LVSum benchmark. Higher values indicate better performance for Kendall’s $\tau$, Spearman’s $\rho$, Content Relevance (CR), and Modality Coherence (MC). The Summary Length column reports the average percentage of video duration selected by each method. V=Video, T=Transcript}
  \label{tab:lvsum-benchmark}
  \begin{center}
    \begin{small}
      \begin{sc}
      \setlength{\tabcolsep}{4.5pt} 
        \begin{tabular}{lcccccr}
          \toprule
          Model  & Model Input & $\tau$& $\rho$ & CR & MC & \makecell{Sum.\\Len.} \\
          \midrule
          Human & - &  0.13388 & 0.13635 & 2.77 & 3.88 & $\approx10\%$ \\\hline
          Random & - & 0.01310 & 0.01357 & 2.14 & 3.91 & $\approx18\%$ \\
          Even & - & 0.00258 & 0.00263 & 2.30 & 3.87 & $\approx18\%$ \\\hline
          Opus-4.5 & V + T & 0.10093 & 0.10513 & 3.03 & 3.90 & $\approx30\%$ \\
          Gemini-2.5-Pro & V + T & 0.08902 & 0.09071 & 3.28 & 4.32 & $\approx12\%$ \\
          Qwen3-VL-235B & V + T & 0.06389 & 0.06706 & 2.40 & 3.90 & $\approx56\%$ \\
          \bottomrule
        \end{tabular}
      \end{sc}
    \end{small}
  \end{center}
\end{table*}
\subsection{Length-Constrained Summary Analysis}
To evaluate model performance under a hard summary-length constraint, we post-process over-long predictions by retaining the highest-ranked segments until the budget is met, adding lower-ranked segments in decreasing importance if needed. Figure~\ref{fig:summary-length-correlation} plots Kendall’s $\tau$ and Spearman’s $\rho$ as a function of enforced summary length. Opus-4.5 shows steady improvements in both $\tau$ and $\rho$ as the budget increases from 10\% to 30\%, reflecting stronger temporal alignment with broader coverage. Gemini-2.5-Pro performs strongly already at 10–15\%, indicating stable salience ranking under compression. In contrast, Qwen3-VL-235B exhibits a non-monotonic trend, dropping around 15\% and remaining substantially below the other models across budgets. At the 15\% budget, Opus-4.5  slightly outperforms Gemini-2.5-Pro, while Qwen3-VL-235B trails significantly. Overall, increasing summary length does not uniformly improve performance and gains depend on model's ability to effectively rank summary segments under compression. Crucially, this compression-conditioned analysis is enabled by LVSum’s interval-level importance annotations with multiple references. Unlike VideoXum and Instruct-V2Xum (Table~\ref{tab:lvsum-benchmark-comparison}), which lack explicit importance ranking, LVSum can assess whether models retain the right events when forced to compress, highlighting its unique value for long-video summarization.


\pgfplotsset{
    corrplot/.style={
        width=0.7\columnwidth,        
        height=0.55\columnwidth,      
        xlabel={Summary Length (\%)},
        ylabel={Correlation},
        ymin=0.04, ymax=0.12,         
        ytick={0.04,0.06,0.08,0.10,0.12},  
        grid=major,
        grid style={dashed, gray!30},
        scaled y ticks=false,
        yticklabel style={
            /pgf/number format/fixed,
            /pgf/number format/precision=2  
        },
        tick align=outside,
        tick label style={font=\tiny},
        label style={font=\tiny},
        legend style={
            font=\tiny,
            draw=none,
            at={(0.98,0.5)},           
            anchor=east,                
            cells={anchor=west},         
            /tikz/every even column/.append style={column sep=4pt}
        },
        legend columns=1,
        mark size=1.2pt
    }
}

\begin{figure}[t]
\centering
\begin{tikzpicture}
\begin{axis}[
    corrplot
]

\addplot+[mark=o, thick]
coordinates {
    (8.71,0.07786)
    (13.51,0.09234)
    (18.14,0.09732)
    (22.73,0.09976)
    (26.42,0.10166)
    (28.73,0.10154)
    (29.70,0.10066)
    (30.02,0.1005)
};
\addlegendentry{Opus-4.5 ($\tau$)}

\addplot+[mark=o, thick, dashed]
coordinates {
    (8.71,0.07881)
    (13.51,0.09369)
    (18.14,0.09957)
    (22.73,0.10299)
    (26.42,0.10553)
    (28.73,0.10569)
    (29.70,0.10482)
    (30.02,0.10468)
};
\addlegendentry{Opus-4.5 ($\rho$)}

\addplot+[mark=square*, thick]
coordinates {
    (8.25,0.0885)
    (11.85,0.08735)
    (12.23,0.08958)
    (12.29,0.08954)
};
\addlegendentry{Gemini-2.5-Pro ($\tau$)}

\addplot+[mark=square*, thick, dashed]
coordinates {
    (8.25,0.08954)
    (11.85,0.089)
    (12.23,0.09129)
    (12.29,0.09124)
};
\addlegendentry{Gemini-2.5-Pro ($\rho$)}

\addplot+[mark=triangle*, thick]
coordinates {
    (8.92,0.05952)
    (14,0.04405)
    (18.52,0.05085)
    (23.14,0.04818)
    (27.73,0.05856)
    (31.60,0.05772)
};
\addlegendentry{Qwen3-VL-235B ($\tau$)}

\addplot+[mark=triangle*, thick, dashed]
coordinates {
    (8.92,0.06009)
    (14,0.04467)
    (18.52,0.05181)
    (23.14,0.04953)
    (27.73,0.06035)
    (31.60,0.05966)
};
\addlegendentry{Qwen3-VL-235B ($\rho$)}

\addplot[
    black,
    dashed,
    thick
] coordinates {(15,0.04) (15,0.12)};

\end{axis}
\end{tikzpicture}


\caption{Correlation vs. summary length for different models. Solid lines denote Kendall's $\tau$, dashed lines denote Spearman's $\rho$.}
\label{fig:summary-length-correlation}
\end{figure}

\subsection{Ablation Analysis}
\subsubsection{Input Modality Ablations}
\label{subsec:input_modality_ablations}
Table~\ref{lvsum-input-modality-ablation} provides insights into how different modalities contribute to MLLM performance. In the transcript-only setup, each model receives only the time-stamped transcript; in the video-only setup, each receives 96 uniformly sampled frames (prompts in \hyperlink{appendix:prompt_video_summarization}{Appendix~\ref*{appendix:prompt_video_summarization}}). Across all three models, the transcript-only (T) modality outperforms video-only (V). For Opus-4.5, moving from V to T yields a 44\% increase in $\tau$ (from 0.06182 to 0.08879), indicating that semantic content from transcripts is the primary driver of ranking and summary relevance. Both Gemini-2.5-Pro and Opus-4.5 exhibit a clear multimodal gain when combining video and transcript (V+T), suggesting that the visual stream contributes complementary non-verbal information absent from transcripts (e.g., on-screen text, demonstrations, emotional cues). A notable finding from the modality ablation is that Qwen3-VL-235B exhibits negative multimodal synergy: its multimodal (V+T) performance drops below T-only input ($\tau$ = 0.04629  vs. 0.07280 vs. 0.06389). This indicates that adding visual tokens actively interferes with the model's ability to rank events, suggesting \textit{\textbf{modality interference}} rather than modality fusion. Interestingly, the T-only condition outperforms V-only, implying that Qwen3-VL-235B primarily treats summarization as a linguistic extraction task. The collapse of performance in the V+T condition coincides with extreme verbosity ($\approx56\%$ summary length), pointing to an \textit{\textbf{instruction drift}} phenomenon where the model disregards global output constraints when processing dense multimodal tokens. The prompt-structure ablation (See \hyperlink{appendix:prompt_positioning}{Appendix~\ref*{appendix:prompt_positioning}}) further supports this: re-injecting the length constraint at the end of the prompt yields the best $\tau/\rho$ and reduces verbosity ($\approx45\%$). However, none of the prompt modifications bring Qwen3-VL-235B close to the intended 15\% budget. These results confirm that it struggles with multimodal fusion and global constraint adherence, and that summarization for this model is bottlenecked more by control and compression than by semantic understanding.

\begin{table}[t]
  \caption{Ablations of input modality. V=Video, T=Transcript}
  \label{lvsum-input-modality-ablation}
  \begin{center}
    \begin{small}
      \begin{sc}
      \setlength{\tabcolsep}{9pt}   
        \begin{tabular}{lcccr}
          \toprule
          Model & Input  & $\tau$  &  $\rho$     & \makecell{Sum.\\ Len.}   \\
          \midrule
          \multirow{3}{*}{Opus-4.5} & V & 0.06182  & 0.06425 & $\approx27\%$  \\
           & T & 0.08879  & 0.09259 & $\approx27\%$   \\
           & V + T & 0.10093 & 0.10513 & $\approx42\%$ \\\hline
          
          \multirow{3}{*}{Gemini-2.5-Pro} & V & 0.06106  & 0.06252 & $\approx14\%$ \\
           & T & 0.07742  & 0.07904 & $\approx12\%$  \\
           & V + T & 0.08902 & 0.09071 & $\approx13\%$ \\\hline
          \multirow{3}{*}{Qwen3-VL-235B} & V & 0.04629  & 0.04958 & $\approx31\%$  \\
           & T & 0.07280  & 0.07594 & $\approx46\%$  \\
           & V + T & 0.06389 & 0.06706 & $\approx56\%$ \\
          \bottomrule
        \end{tabular}
      \end{sc}
    \end{small}
  \end{center}
\end{table}

\subsubsection{Video Sampling Density Ablations}
\begin{table}[t]
  \caption{Ablations on number of input frames for video-only evaluation}
  \label{tab:lvsum-video-frame-sampling-ablation}
  \begin{center}
    \begin{small}
      \begin{sc}
        \setlength{\tabcolsep}{6pt}
        \begin{tabular}{lcccr}
          \toprule
          Model & \# Frames  & $\tau$  & $\rho$ & \makecell{Sum.\\ Len.} \\
          \midrule
          \multirow{5}{*}{Gemini-2.5-Pro} & 24 & 0.03645 & 0.03740 &$\approx14\%$ \\
           & 32 & 0.04221 &0.04298 &$\approx12\%$ \\
           & 64 &0.05580&0.05721&$\approx14\%$ \\
           & 96 &0.05580&0.05714&$\approx14\%$ \\
           & 128 &0.05734&0.05854 &$\approx12\%$ \\ \hline
          \multirow{5}{*}{Qwen3-VL-235B} & 24 & 0.03559 & 0.03674 &$\approx25\%$ \\
           & 32 & 0.04184 & 0.04307 &$\approx27\%$ \\
           & 64 &0.04796&0.04958&$\approx31\%$ \\
           & 96 &0.04629&0.04774&$\approx34\%$ \\
          & 128 & 0.04876 & 0.05020 &$\approx32\%$ \\

          \bottomrule
        \end{tabular}
      \end{sc}
    \end{small}
  \end{center}
\end{table}


Table~\ref{tab:lvsum-video-frame-sampling-ablation} presents an ablation on temporal granularity, measured by the number of sampled input frames, for video-only summarization using Gemini-2.5-Pro and Qwen3-VL-235B. Both models show an overall improvement in ranking accuracy ($\tau$) as frame count increases, peaking at 128 frames (Gemini-2.5-Pro: $\tau=0.05734$; Qwen3-VL-235B: $\tau=0.04876$). Interestingly, Qwen3-VL-235B exhibits a dip at 96 frames, suggesting sensitivity to specific temporal configurations, potentially reflecting architectural or training alignment with power-of-two frame counts (e.g., 64 and 128). Overall, increased temporal context consistently benefits ranking performance, with Gemini-2.5-Pro showing stronger gains while maintaining concise summaries (10–14\%). Another key observation that emerges from comparing Table~\ref{tab:lvsum-video-frame-sampling-ablation} and Table~\ref{lvsum-input-modality-ablation} is that it provides additional evidence supporting the \textit{\textbf{modality interference}} argument presented in Section~\ref{subsec:input_modality_ablations}.  In Table~\ref{tab:lvsum-video-frame-sampling-ablation}, Qwen3-VL-235B handles a high visual load (128 frames) while maintaining 32\% summary length.  However, in Table~\ref{lvsum-input-modality-ablation}, when we combine 96 frames with a Transcript (V+T), the length explodes to 56\%. This proves that the ``forgetting'' of length constraints in Qwen3-VL-235B is not caused by the number of tokens alone (as it handles 128 video frames fine), but specifically by the interference between visual and textual modalities when they are interleaved.

\begin{table*}[t]
  \caption{Ablations by category on Video + Transcript input}
  \label{tab:lvsum-category-wise-ablation}
  \begin{center}
    \begin{small}
      \begin{sc}
      \setlength{\tabcolsep}{4.5pt}
        \begin{tabular}{lcccccr}
          \toprule
          Model & Category  &  $\tau$      & $\rho$ & CR & MC & \makecell{Sum.\\ Len.} \\
          \midrule
          \multirow{5}{*}{Opus-4.5} & Lectures   & 0.11163  & 0.11633 & 3.22  & 4.21 & $\approx31\%$ \\
           & Podcasts   & 0.10354  & 0.10820 & 2.94  & 4.13 & $\approx30\%$  \\
           & Vlogs  & 0.10852  & 0.11220 & 3.05 &  3.00 &  $\approx24\%$\\
           & Explainer Videos  & 0.10855 & 0.11340  & 2.90 & 3.75 & $\approx31\%$\\
           & News Reports  & 0.07502 & 0.07813  & 2.77 & 3.75 & $\approx33\%$\\\hline
          \multirow{5}{*}{Gemini-2.5-Pro} & Lectures   & 0.08215  & 0.08399 &  3.58 & 4.46 & $\approx13\%$ \\
           & Podcasts   & 0.09564  & 0.09767 & 3.15 & 4.63 & $\approx13\%$ \\
           & Vlogs  & 0.13420  & 0.13666 & 3.45  & 4.00 & $\approx14\%$   \\
           & Explainer Videos  & 0.07776 & 0.07920 & 3.64 & 4.25 & $\approx15\%$\\
           & News Reports  & 0.11468 & 0.11627 &  3.21 & 5.00 & $\approx12\%$\\\hline
          \multirow{5}{*}{Qwen3-VL-235B} & Lectures   & 0.06938  & 0.07310 & 2.51 & 4.18 & $\approx57\%$ \\
           & Podcasts   & 0.07896  & 0.08281 & 2.41  & 4.25 & $\approx60\%$ \\
           & Vlogs  & 0.09249  & 0.09671 & 2.01  & 3.00 & $\approx62\%$ \\
           & Explainer Videos  & 0.00901 & 0.00931 & 2.49  & 3.00 & $\approx60\%$ \\
           & News Reports  & 0.02433 & 0.02555 & 2.41  & 4.25 &  $\approx55\%$ \\

          \bottomrule
        \end{tabular}
      \end{sc}
    \end{small}
  \end{center}
\end{table*}

\subsubsection{Ablations by Categories}


Category-wise analysis (Table ~\ref{tab:lvsum-category-wise-ablation}) of LVSum's five largest domains reveals that long-video summarization difficulty is strongly content-dependent, with models exhibiting distinct inductive biases. Gemini-2.5-Pro shows strong temporal agreement on Vlogs and News Reports while adhering to its budget and achieving high modality coherence, indicating robust cross-modal grounding. Opus-4.5 excels in salience ranking for structured domains like Lectures but systematically exceeds the length budget, reflecting a tendency towards semantic extraction. Conversely, Qwen3-VL-235B exhibits pronounced category disparity, performing moderately on conversational formats but collapsing on event-dense domains while remaining highly verbose. These findings suggest that linguistic scaffolding strongly predicts ranking performance, whereas compression and multimodal grounding are model-dependent behaviors, and category-level evaluation exposes systematic biases that global averages often obscure.

\subsection{Qualitative Analysis}
\subsubsection{Comparison of outputs from various models}

A qualitative comparison (Figure~\ref{fig:single-video-comparison}) illustrates how different models summarize the same long video and highlights gaps in current MLLM-based summarization. The human summary is concise and event-centric, selecting key narrative beats (e.g., the performance, the historic studio context, the drum used by artists, and the producer's closing remarks), while omitting peripheral details. Opus-4.5 captures most salient events and preserves narrative ordering, but produces a longer and more descriptive summary than necessary, reflecting a tendency toward semantic extraction rather than compression. Gemini-2.5-Pro produces video summaries by selecting a compact set of segments that adhere to the main storyline with minimal redundancy, consistent with its superior budget compliance observed quantitatively. In contrast, Qwen3-VL-235B includes nearly all events in the video, yielding a verbose summary that resembles a paraphrased transcript rather than a compressed key-shot summary. While Qwen3-VL-235B shows strong semantic understanding, it fails to distinguish primary from secondary content and frequently extends beyond the intended summary boundary, corroborating the \textit{\textit{instruction drift}} and verbosity effects observed in modality and constraint analyses. Overall, the qualitative results reinforce that long-video summarization requires not only semantic comprehension but also temporal prioritization and compression—capabilities that current MLLMs inconsistently exhibit.

To further illustrate the generalization of these patterns across different video types, we did a side-by-side comparison of model summaries for two more videos with distinct content: a news report about monsoon flooding in Mumbai and a school trip video to Normandy (See Appendix ~\ref{appendix:ablations_frames_opus}). Across both videos, we observe consistent behavior: Opus-4.5 provides detailed but lengthy summaries, Gemini-2.5-Pro produces concise summaries that align well with human selections, and Qwen3-VL-235B generates verbose outputs that include most video events regardless of their relative importance. This cross-video consistency reinforces that the summarization patterns observed in Figure~\ref{fig:single-video-comparison} reflect systematic model behavior rather than video-specific artifacts.

\subsubsection{Comparison of MLLM-as-Judge metrics}
\begin{figure*}[!ht]
\centering
\includegraphics[width=1.0\linewidth]{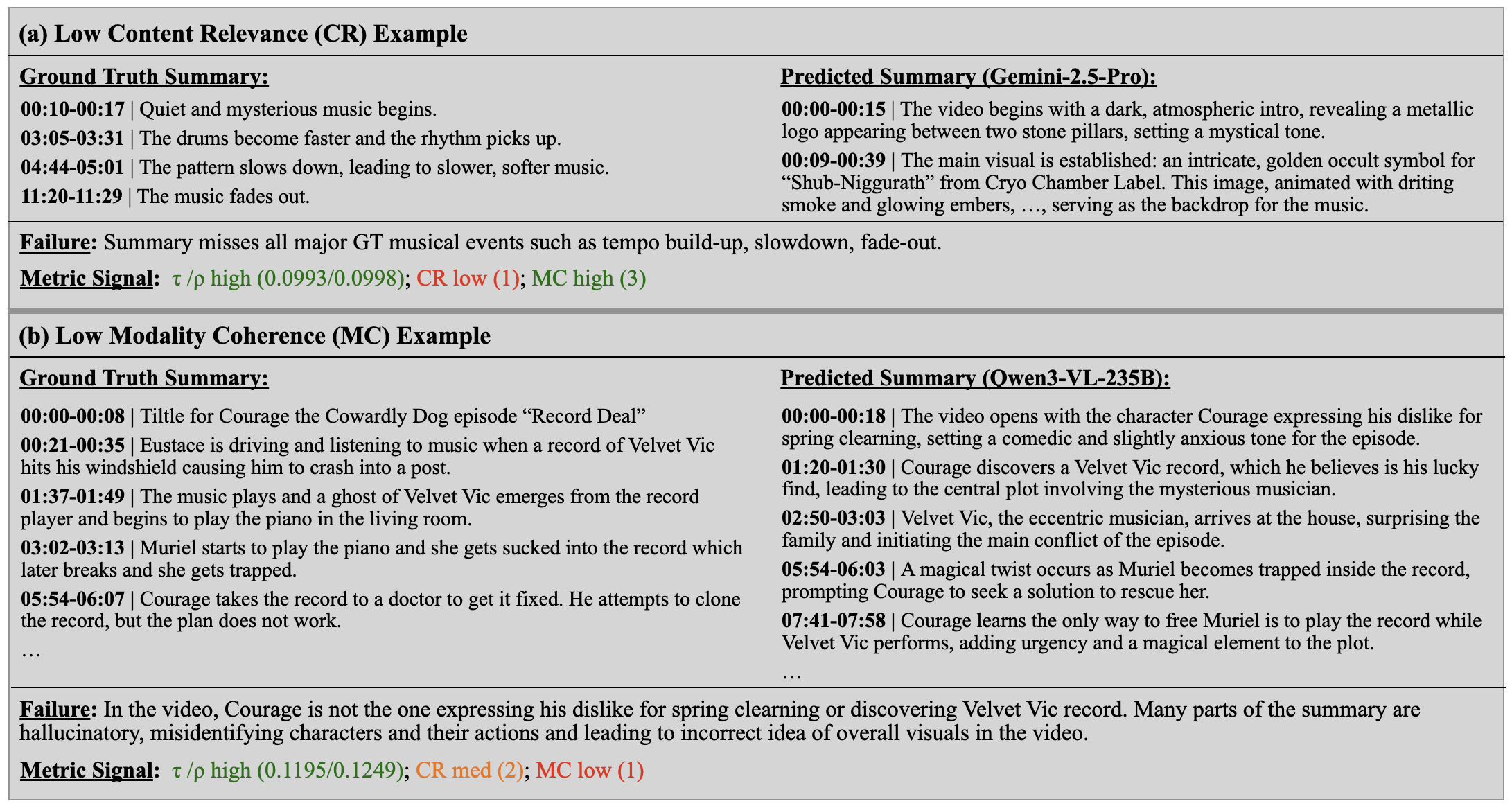}
\caption{Failure cases illustrating distinct evaluation modes. (a) Low Content Relevance (CR): summary omits salient events. (b) Low Modality Coherence (MC): textual descriptions contradict visual evidence within the predicted interval.}
\label{fig:failure_modes}
\end{figure*}

While rank-based metrics such as $\tau$ and $\rho$ capture temporal importance alignment, they do not evaluate whether the predicted summary contains the right content or whether the description is grounded in the corresponding visual/audio evidence. The failure cases in Figure~\ref{fig:failure_modes} make this distinction explicit.
\input{single_video_comparison_figure_v2.tex}
\FloatBarrier
 In the Low Content Relevance (CR) example, the model (Gemini-2.5-Pro) produces a well-formed and temporally organized summary but omits the primary narrative beats, resulting in a semantically misaligned summary despite reasonable temporal ranking and modality grounding. Conversely, the Low Modality Coherence (MC) example shows the opposite failure mode: the model (Qwen3-VL-235B) selects semantically appropriate events, yet the characters and their actions in the textual description are not aligned with the predicted video intervals, leading to poor modality grounding.  Notably, both failure types would be scored high by $\tau/\rho$ metrics. CR and MC therefore quantify two complementary dimensions of summarization quality—semantic relevance and cross-modal grounding—that are essential for realistic, MLLM-based video summarization.


\FloatBarrier

\section{Conclusion}
In this work we introduced LVSum, a benchmark for timestamp-aware long-video summarization with multi-reference human annotations and interval-level importance supervision. Through comprehensive evaluation of both proprietary and open-source MLLMs, we showed that current systems exhibit strong semantic understanding yet lack robust temporal compression, multimodal grounding, and constraint adherence. Our analyses further exposed category-dependent behavior and failure modes that are obscured under traditional rank-based metrics, motivating the need for richer evaluation dimensions such as \textbf{Content Relevance} and \textbf{Modality Coherence}. We hope LVSum provides a foundation for studying temporally-grounded summarization and stimulates research toward MLLMs that balance semantic salience, timestamp precision, and multimodal consistency for real-world long-video understanding.

\bibliographystyle{splncs04}
\bibliography{main}

\newpage
\appendix
\hypertarget{appendix:annotation_guidelines}{}
\section{Annotation Guidelines for LVSum Dataset}
\label{appendix:annotation_guidelines}
\begin{mdframed}[
    linewidth=1pt,
    linecolor=black,
    backgroundcolor=gray!5,
    nobreak=true,        
    innertopmargin=8pt,
    innerbottommargin=8pt
]

\small
\noindent\textbf{Goal}: To create a short, informative summary of each video by selecting, describing, and ranking the most relevant segments.

\noindent\textbf{General Guidance}:
\begin{itemize}[noitemsep, topsep=0pt]
    \item \textit{Relevance}: Focus on segments that are most relevant to the core topic and purpose of the video.
    \item \textit{Accuracy}: The selected segments and their descriptions must accurately reflect the content of the video.
    \item \textit{Conciseness}: The summary should be as short as possible while still conveying the key information.
    \item \textit{Non-Redundancy}: Avoid including segments that repeat information already covered.
    \item \textit{Informativeness}: Prioritize segments that provide new or important information.
\end{itemize}

\subsection*{Video Segmentation}
This step involves segmenting the video to identify key moments that reflect its main purpose, enabling a concise summary of its core content. Ask yourself: "What are the most crucial moments in this video?" Your segment choices should answer that question.

\noindent\textbf{DO Select the Segments}:
\begin{itemize}[nosep]
    \item \textit{Informative}: Choose segments that convey meaningful information related to video.
    \item \textit{Major Highlights}: Prioritize segments showcasing the most exciting, impactful, or informative moments.
    \item \textit{Fact-Based Representation}: The segment must reflect the factual audio and visual content. Avoid inferences.
\end{itemize}

\noindent\textbf{DO NOT Select the Segments}:
\begin{itemize}[nosep]
    \item \textit{Low-Value}: Waiting (loading screens), Silence (no commentary), or Transitions.
    \item \textit{Overlapping}: Avoid segments that duplicate information presented elsewhere.
    \item \textit{Irrelevant}: Avoid segments unrelated to the video's primary topic.
\end{itemize}

\noindent\textbf{Segment Length Restriction}: Ensure the segment fully covers the target information.
\begin{itemize}[nosep]
    \item \textbf{Maximum Length}: Total duration must NOT exceed 15\% of total video length.
    \item \textit{Example}: 5-min video = 45s max; 10-min video = 90s max.
\end{itemize}
\end{mdframed}
\clearpage
\begin{mdframed}[
    linewidth=1pt,
    linecolor=black,
    backgroundcolor=gray!5,
    nobreak=true,        
    innertopmargin=8pt,
    innerbottommargin=8pt
]
\subsection*{Segment Ranking}
This step involves ranking segments by relevance. Ask yourself: "How essential is each segment to understanding the video's main message?"
\FloatBarrier
\begin{table}[H]
\centering
\caption{Ranking Scale for Video Segment Importance}
\label{tab:ranking-scale}
\begin{tabularx}{\textwidth}{c >{\bfseries}l X}
\toprule
\textbf{Scale} & \textbf{Definition} & \textbf{Description} \\ \midrule
3 & Highly Relevant (Essential) & Crucial for understanding. Removing it significantly detracts from the summary. Represents a key highlight. \\ \midrule
2 & Relevant (Important) & Provides valuable context. Not essential, but inclusion strengthens understanding. \\ \midrule
1 & Somewhat Relevant & Less critical. Adds value but could be omitted without impacting the core message. Use only if time remains in 15\% limit. \\ \bottomrule
\end{tabularx}
\end{table}

\subsection*{Review Your Task}
A final review ensures the most essential segments are prioritized and descriptions are accurate.

\noindent\textbf{Review Checklist}:
\begin{itemize}[nosep]
    \item \textbf{Re-Rank}: Ensure the most crucial segments have a rank of 3.
    \item \textbf{Refine}: Ensure descriptions are accurate, concise, and consistent.
    \item \textbf{Final Check}: Total length $\le$ 15\%; no overlaps; all segments ranked 1--3.
\end{itemize}
\end{mdframed}
\clearpage
\hypertarget{appendix:prompt_video_summarization}{}
\section{Prompts for Video Summarization}
\label{appendix:prompt_video_summarization}
\hypertarget{appendix:transcript-only-prompt}{}
\subsection{Transcript-Only Prompt}
\label{appendix:transcript-only-prompt}
\begin{mdframed}[linewidth=1pt, linecolor=black, backgroundcolor=gray!5]
\small
You are given a video transcript. Please provide a concise summary consisting of multiple important segments extracted from the video transcripts. For each segment, include:

The start and end timestamps in the format HH:MM:SS - HH:MM:SS.
A relevance score from 0 to 3, where 0 means least important and 3 means most important.
A brief textual description of the segment content.

Current video length is \texttt{\{video\_length\}} seconds. The combined duration of all summarized segments should not exceed 15\% of the total length of the video.

Transcript:
\texttt{\{transcript\}}

Format your output as a list, for example:

Segment 1: 00:00:10 - 00:00:40 | Score: 3 | Description: Introduction to the main topic.\\
Segment 2: 00:05:20 - 00:05:50 | Score: 2 | Description: Explanation of key concept.\\
Segment 3: 00:12:00 - 00:12:30 | Score: 1 | Description: Additional example provided.
\end{mdframed}

\clearpage
\hypertarget{appendix:video-plus-transcript-prompt}{}
\subsection{Video-Only and Video+Transcript Prompt}
\label{appendix:video-plus-transcript-prompt}

\begin{mdframed}[linewidth=1pt, linecolor=black, backgroundcolor=gray!5]
\small
You are given a video. Please provide a concise summary consisting of multiple important segments extracted from the video. For each segment, include:

The start and end timestamps in the format HH:MM:SS - HH:MM:SS.
A relevance score from 0 to 3, where 0 means least important and 3 means most important.
A brief textual description of the segment content.

Current video length is \texttt{\{video\_length\}} seconds. The combined duration of all summarized segments should not exceed 15\% of the total length of the video.

Here are the frames that I will upload and transcripts. These frames are sampled at \texttt{\{timestamp\_list\_in\_HH\_MM\_SS\}} \\
Transcript:

\texttt{\{transcript\}}

Generate a video summary using both the visual content in the video frames and the transcript (if provided).

Format your output as a list, for example:\\
Segment 1: 00:00:10 - 00:00:40 | Score: 3 | Description: Introduction to the main topic.\\
Segment 2: 00:05:20 - 00:05:50 | Score: 2 | Description: Explanation of key concept.\\
Segment 3: 00:12:00 - 00:12:30 | Score: 1 | Description: Additional example provided.

\end{mdframed}

When transcript is not available in Video-only model, we use transcript="[No transcript provided]" in above prompt.

\clearpage
\hypertarget{appendix:mllm-as-judge}{}
\section{Prompts for MLLM-as-Judge}
\label{appendix:mllm-as-judge}
\hypertarget{appendix:content-relevance-prompt}{}
\subsection{Content Relevance Prompt}
\label{appendix:content-relevance-prompt}

\begin{mdframed}[linewidth=1pt, linecolor=black, backgroundcolor=gray!5]
\small
You are an expert video-summarization evaluator.

Your task is to assess **Content Relevance** between a Generated Summary and a Ground-Truth Summary.

Definition:\\
- Content Relevance measures how well the generated summary captures the important events, actions, objects, and semantic information from the ground truth.\\
- Focus on semantic correctness, event coverage, and whether essential information is preserved.\\
- Do NOT penalize for stylistic or wording differences.\\
- Penalize missing key events, adding irrelevant information, or misrepresenting the main ideas.

Given:\\
\texttt{[GROUND\_TRUTH\_SUMMARY]}\\
\texttt{\{ground\_truth\_summary\}}

\texttt{[GENERATED\_SUMMARY]}\\
\texttt{\{predicted\_summary\}}

Evaluation criteria:\\
1. **Key Event Coverage** – Are the important events from GT present?\\
2. **Semantic Accuracy** – Are the actions, objects, outcomes correctly preserved?\\
3. **Irrelevant Content** – Does the generated summary add unnecessary or incorrect info?\\
4. **Overall Alignment** – Does the generated summary convey the same meaning as GT?

Provide:\\
- A score from **1 to 5** (integer only).\\
  1 = Very poor relevance; 5 = Excellent alignment with ground truth.\\
- One short justification.

Output format (strict):\\
Score: X\\
Justification: <one sentence>
\end{mdframed}

\clearpage
\hypertarget{appendix:modality-coherence-prompt}{}
\subsection{Modality Coherence Prompt}
\label{appendix:modality-coherence-prompt}

\begin{mdframed}[linewidth=1pt, linecolor=black, backgroundcolor=gray!5]
\small
You are an expert evaluator for multimodal grounding.

Your task is to assess **Modality Coherence** between a Generated Summary and the visual/audio content of the video.

Definition:
- Modality Coherence measures whether each statement in the generated summary is supported by what is present in the video frames or audio.\\
- Check that objects, actions, attributes, sounds, and scenes mentioned in the summary actually appear in the video.\\
- Penalize hallucinations, contradictions, and modality mismatches (visual or auditory).

Given:\\
\texttt{[VIDEO\_VISUAL\_DESCRIPTIONS]}\\
\texttt{\{video\_descriptions\}}

\texttt{[VIDEO\_AUDIO\_TRANSCRIPT]} (if any)\\
\texttt{\{audio\_transcripts\}}

\texttt{[GENERATED\_SUMMARY]}\\
\texttt{\{predicted\_summary\}}

Evaluation criteria:\\
1. **Visual Grounding** – Do mentioned objects/actions exist in the frames?\\
2. **Audio-Visual Consistency** – Are sound-related statements supported by audio?\\
3. **Hallucination Check** – Does the summary invent objects/actions/scenes?\\
4. **Cross-Modal Agreement** – Are descriptions mutually consistent across modalities?

Provide:\\
- A score from **1 to 5** (integer only).\\
  1 = many hallucinations; 5 = fully grounded and consistent summary.\\
- One sentence of justification.

Output format (strict):\\
Score: X\\
Justification: <one sentence>
\end{mdframed}

\clearpage
\hypertarget{appendix:prompt_positioning}{}
\section{Constraint Compliance Test for Qwen3VL}
\label{appendix:prompt_positioning}
To verify if Qwen3-VL-235B is indeed forgetting the constraint due to input length in Video (V) + Transcript (T) mode, we ran  ablation experiment with following prompt re-positioning:
\begin{itemize}
    \item \textbf{A Length Constraint First)}: [Length Constraint] + [Transcript] + [Video]
    \item \textbf{B (Length Constraint Last)}: [Transcript] + [Video] + [Length Constraint]
    \item \textbf{C (Sandwich)}: [Length Constraint] + [Transcript] + [Video] + [Length Constraint]
\end{itemize}

The intuition is that if the model complies better in Condition B or C, it proves a "Recency Bias" or "Position Forgetting" in V+T mode of model evaluation.

\begin{table}[htbp]
  \caption{Ablations by category, Qwen3-VL-235B on Video + Transcript input}
  \label{lvsum-constraints}
  \begin{center}
    \begin{small}
      \begin{sc}
        \begin{tabular}{lccr}
          \toprule
          Model  &  $\tau$  & $\rho$  & Sum. Len. \\
          \midrule
          O (In this paper) & 0.06389 & 0.06706 & $\approx56\%$\\
          A (Length Constraint First) & 0.06389 & 0.06698 & $\approx58\%$ \\
          B (Length Constraint Last)   &  0.07396 & 0.07703 & $\approx45\%$  \\
          C (Sandwich)  & 0.07198  & 0.07549 & $\approx50\%$\\
          \bottomrule
        \end{tabular}
      \end{sc}
    \end{small}
  \end{center}
\end{table}

\clearpage
\hypertarget{appendix:additional_examples}{}
\section{Additional Qualitative Examples}
\label{appendix:additional_examples}

\input{multi_video_comparison_table.tex}

\clearpage
\hypertarget{appendix:ablations_frames_opus}{}
\section{Additional Ablations on number of input frames for video-only evaluations with Opus-4.5 model}
\label{appendix:ablations_frames_opus}
\begin{table}[htbp]
  \caption{Ablations on number of input frames for video-only evaluation}
  \label{lvsum-video-frame-sampling-ablation}
  \begin{center}
    \begin{small}
      \begin{sc}
        \setlength{\tabcolsep}{6pt}
        \begin{tabular}{lcccr}
          \toprule
          Model & \# Frames  & $\tau$  & $\rho$ & \makecell{Sum.\\ Len.} \\
          \midrule
          \multirow{5}{*}{Opus-4.5} & 24 & 0.04582 & 0.04759 &$\approx27\%$ \\
           & 32 & 0.06170 &0.06408 &$\approx26\%$ \\
           & 64 &0.05404&0.05614&$\approx28\%$ \\
           & 96 &0.06329&0.06569&$\approx26\%$ \\
           & 100 &0.06401&0.06639 &$\approx25\%$ \\ \hline
          \bottomrule
        \end{tabular}
      \end{sc}
    \end{small}
  \end{center}
\end{table}

\end{document}

%% file: single_video_comparison_figure_v2.tex

\providecommand{\thickhline}{\noalign{\hrule height 1.2pt}}
\newlength{\mytablewidth}

\begin{figure*}[t]
\centering
\caption{Video summarization comparison for a selected video from LVSum.}
\label{fig:single-video-comparison}

\setlength{\tabcolsep}{6pt}
\renewcommand{\arraystretch}{1.0}
\setlength{\arrayrulewidth}{0.6pt}
\setlength{\mytablewidth}{\dimexpr\linewidth - 2\tabcolsep - 2\arrayrulewidth\relax}

\begin{tabular}{|p{\mytablewidth}|}
\hline

\begin{minipage}{\mytablewidth}
  \vspace{4pt}
  \centering
  \textbf{Human Summary}
  \vspace{4pt}
\end{minipage}
\\
\hline

\begin{minipage}{\mytablewidth}
  \vspace{4pt}
  \centering
  \includegraphics[width=0.95\mytablewidth,keepaspectratio]{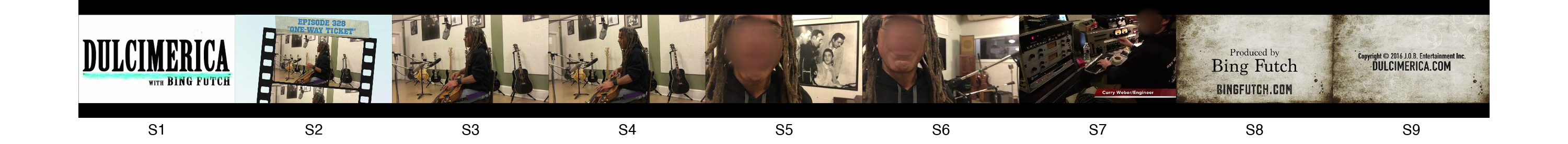}
  \vspace{4pt}
\end{minipage}
\\
\hline

\begin{minipage}{\mytablewidth}
\vspace{2pt}
\tiny
\textbf{S1} (00:03--00:10): score: 3: The title Dulcimeria with Bing Futch is displayed.\\
\textbf{S2} (00:12--00:16): score: 3: The title, Episode 328 -- ``One-way Ticket'' is displayed.\\
\textbf{S3} (00:32--00:40): score: 3: The artist starts singing the song ``One-Way Ticket'' and playing guitar.\\
\textbf{S4} (04:30--04:36): score: 3: The artist completes singing the song ``One-Way Ticket''.\\
\textbf{S5} (05:04--05:12): score: 3: The artist mentions that he is in Sun Studio and completed tracking on an untitled album.\\
\textbf{S6} (05:47--05:52): score: 2: The artist starts giving a tour of the studio.\\
\textbf{S7} (06:56--07:02): score: 3: The artist introduces the engineer, Curry.\\
\textbf{S8} (08:38--08:42): score: 3: End credits start being displayed.\\
\textbf{S9} (09:21--09:23): score: 3: End credits complete.
\vspace{2pt}
\end{minipage}
\\

\thickhline

\begin{minipage}{\mytablewidth}
  \vspace{4pt}
  \centering
  \textbf{Opus-4.5}
  \vspace{4pt}
\end{minipage}
\\
\hline

\begin{minipage}{\mytablewidth}
  \vspace{4pt}
  \centering
  \includegraphics[width=0.95\mytablewidth,keepaspectratio]{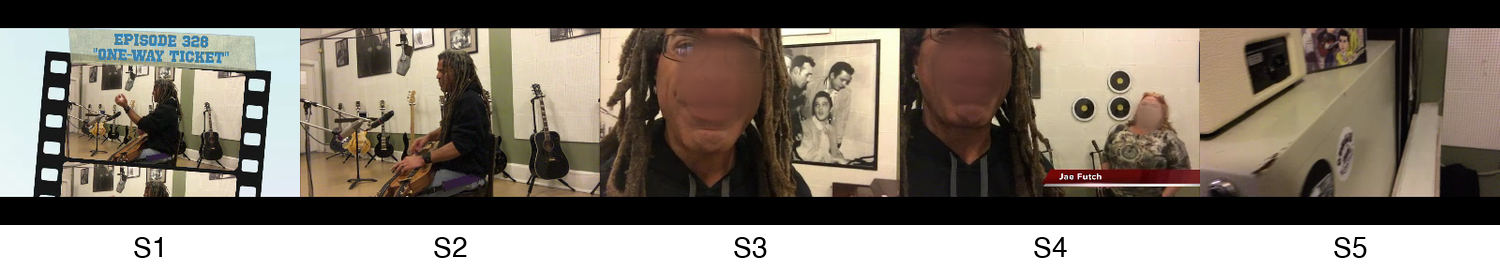}
  \vspace{4pt}
\end{minipage}
\\
\hline

\begin{minipage}{\mytablewidth}
\vspace{2pt}
\tiny

\textbf{S1} (00:06--00:30): score: 2: Introduction showing ``Dulcimerica with Bing Futch'' title card and Episode 328 ``One-Way Ticket'' - establishes the episode context and shows the musician preparing to record at Sun Studio.\\
\textbf{S2} (00:30--01:10): score: 3: The artist performs the opening of ``One-Way Ticket'' on dulcimer at Sun Studio, showcasing the live recording session with vocals and instrumentation in the historic space.\\
\textbf{S3} (05:04--05:47): score: 3: Post-recording commentary where the artist explains he just completed tracking an album at Sun Studio featuring songs he hasn't performed in over 30 years, recorded ``old school'' style in about two hours.\\
\textbf{S4} (05:47--06:22): score: 2: Tour of Sun Studio showing historic photographs on the walls and the famous drum set used by U2's Larry Mullen Jr. during the ``Rattle and Hum'' recording sessions.\\
\textbf{S5} (07:26--08:00): score: 2: Introduction of engineer Curry Weber in the control booth, showing the mix of high-tech and vintage equipment, plus a humorous tour of the bathroom featuring gold records on the walls.

\vspace{2pt}
\end{minipage}
\\

\thickhline

\begin{minipage}{\mytablewidth}
  \vspace{4pt}
  \centering
  \textbf{Gemini-2.5-Pro}
  \vspace{4pt}
\end{minipage}
\\
\hline

\begin{minipage}{\mytablewidth}
  \vspace{4pt}
  \centering
  \includegraphics[width=0.95\mytablewidth,keepaspectratio]{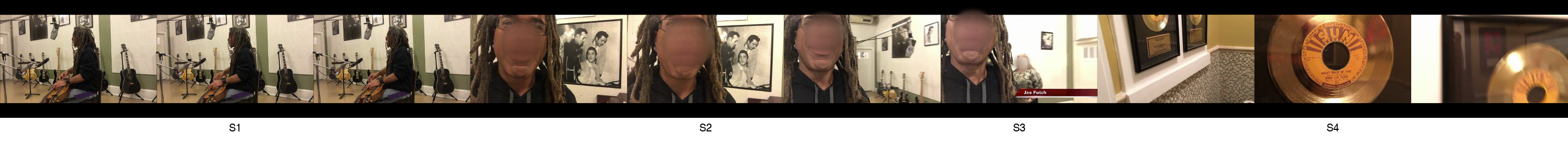}
  \vspace{4pt}
\end{minipage}
\\
\hline

\begin{minipage}{\mytablewidth}
\vspace{2pt}
\tiny
\textbf{S1} (00:42--01:10): score: 3: Bing Futch performs the first verse and chorus of his song ``One-Way Ticket'' on the mountain dulcimer inside the recording studio.\\
\textbf{S2} (05:04--05:25): score: 3: The artist explains that he is at the historic Sun Studio in Memphis, recording an album of old songs ``old school'' style by singing and playing simultaneously.\\
\textbf{S3} (05:47--06:15): score: 2: A brief tour of the famous recording space, highlighting the drum set used by U2's Larry Mullen Jr.\ for the album ``Rattle and Hum''.\\
\textbf{S4} (07:46--08:11): score: 1: A quick and humorous look inside the studio's unique bathroom, which is decorated with gold records and features a guitar-shaped toilet seat.
\vspace{2pt}
\end{minipage}
\\

\thickhline

\begin{minipage}{\mytablewidth}
  \vspace{4pt}
  \centering
  \textbf{Qwen3-VL-235B}
  \vspace{4pt}
\end{minipage}
\\
\hline

\begin{minipage}{\mytablewidth}
  \vspace{4pt}
  \centering
  \includegraphics[width=0.95\mytablewidth,keepaspectratio]{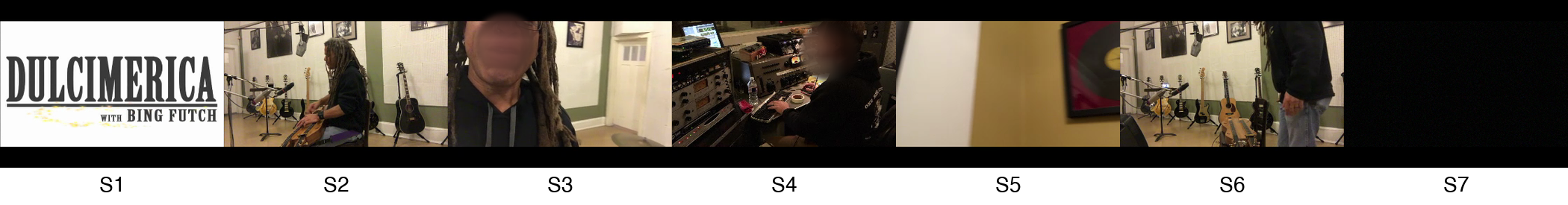}
  \vspace{4pt}
\end{minipage}
\\
\hline

\begin{minipage}{\mytablewidth}
\vspace{2pt}
\tiny
\textbf{S1} (00:00--00:06): score: 1: Opening title card for ``DULCIMERICA with BING FUTCH'' and episode 326 titled ``ONE-WAY TICKET'' displayed over a filmstrip graphic featuring the artist in a studio. \\
\textbf{S2} (00:07--01:45): score: 3: The artist performs the song ``One-Way Ticket'' on a lap steel guitar in a historic recording studio, surrounded by multiple guitars and microphones, with lyrics about leaving a small town and embracing destiny.\\
\textbf{S3} (05:04--06:48): score: 3: The artist speaks to the camera, explaining he just finished recording an album at Sun Studio, highlighting the historic significance of the space and the U2 drum set used for ``Angel of Harlem.''\\
\textbf{S4} (06:48--07:26): score: 2: The artist introduces engineer Curry Weber in the control room, praising his work and the unique blend of high-tech and lo-fi equipment in the studio.\\
\textbf{S5} (07:39--08:22): score: 2: The artist shows the studio's bathroom, which features gold records on the wall and a guitar-shaped toilet seat, expressing amusement and admiration for the quirky decor. \\
\textbf{S6} (09:19--09:39): score: 2: The artist reflects on recording under a photo of Elvis Presley, acknowledging the daunting yet inspiring atmosphere of Sun Studio. \\
\textbf{S7} (09:49--10:19): score: 1: Closing credits and sponsor acknowledgments, including endorsements for V-Picks and Zither Music Company, followed by production credits.

\vspace{2pt}
\end{minipage}
\\
\hline

\end{tabular}
\end{figure*}

%% file: multi_video_comparison_table.tex
\begin{figure*}[!ht]
\centering
\caption{Comparison of video summarization results across different models and videos}
\label{fig:multi-video-comparison}
\scalebox{0.65}{
\begin{tabular}{|p{0.644\textwidth}|p{0.644\textwidth}|}
\hline
\textbf{Video 1} & \textbf{Video 2} \\
\hline
\multicolumn{2}{|c|}{\textbf{Human Grading}} \\
\hline
\includegraphics[width=\linewidth,keepaspectratio]{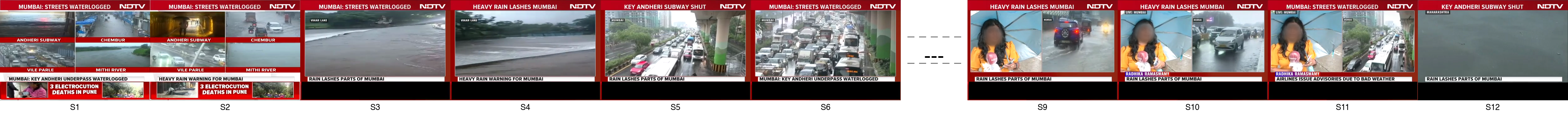} & \includegraphics[width=\linewidth,keepaspectratio]{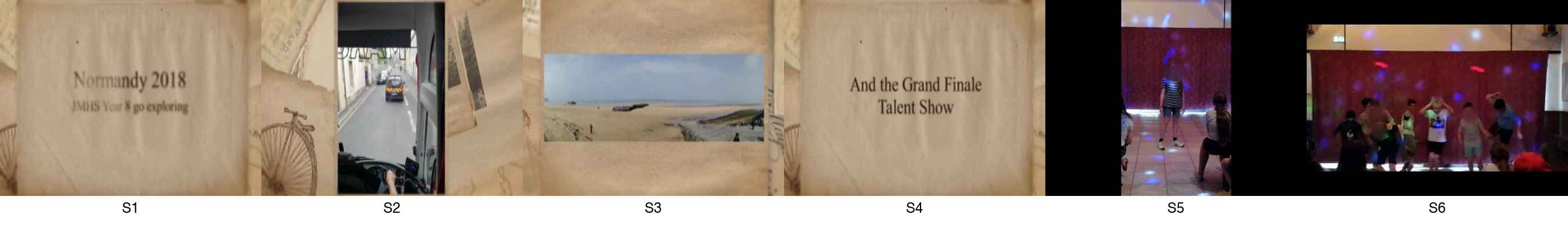} \\
\hline
\begin{minipage}{0.95\linewidth}
\tiny
\textbf{S1} (00:04--00:09): score: 3: Heavy rain lashed Mumbai, caused an overflow of water supplying lake. \\
\textbf{S2} (00:23--00:27): score: 2: Some schools and colleges have been shut down in several districts. \\
\textbf{S3} (00:37--00:41): score: 3: Rainfall is over 60 mm in various spots. \\
\textbf{S4} (00:44--00:49): score: 2: 6 teams has been deployed in India to deal with the consequences of the disaster. \\
\textbf{S5} (01:03--01:08): score: 3: A warning is released regarding possible delays in air travel.\\
\textbf{S6} (01:23--01:26): score: 2: Also the ramping wind with 50-60 km/h is an impactful factor.\\
\textit{---} \\
\textbf{S9} (03:16--03:24): score: 2: Certain parts of western suburbs and eastern suburbs have been severely impacted by the flooding.\\
\textbf{S10} (04:50--05:07): score: 3: South Bombay and certain areas of western suburbs are much less impacted than Andheri, Sion, Kurla, Chembur.\\
\textbf{S11} (06:26--06:36): score: 2: Schools in Palghar, Raigarh etc. have been shut, waiting for further advice from the government.\\
\textbf{S12} (08:05--08:10): score: 2: Improvements of the situations are very inconsistent, as rain stops for several minutes and then comes back.
\end{minipage} & \begin{minipage}{0.95\linewidth}
\tiny
\textbf{S1} (00:00--00:04): score: 3: The \"Normandy 2018 JMHS Year 8\" trip video starts with the title on a beige backdrop. \\
\textbf{S2} (00:29--00:35): score: 2: The students got a police escort.\\
\textbf{S3} (00:51--01:07): score: 2: The students' trip to Gold Beach appears as a transition and photos of the trip are shown.\\
\textbf{S4} (03:48--03:50): score: 3: The grand finale talent show is introduced.\\
\textbf{S5} (05:23--05:28): score: 2: A young girl with red hair delivers her poem to the audience.\\
\textbf{S6} (12:34--12:51): score: 2:A group of boys dance and jump to music nonsensically.
\end{minipage} \\
\noalign{\hrule height 2pt}
\multicolumn{2}{|c|}{\textbf{Opus 4.5}} \\
\hline
\includegraphics[width=\linewidth,keepaspectratio]{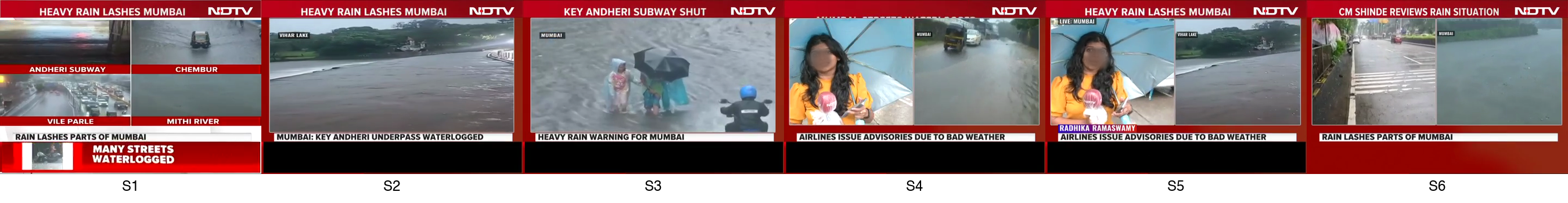} & \includegraphics[width=\linewidth,keepaspectratio]{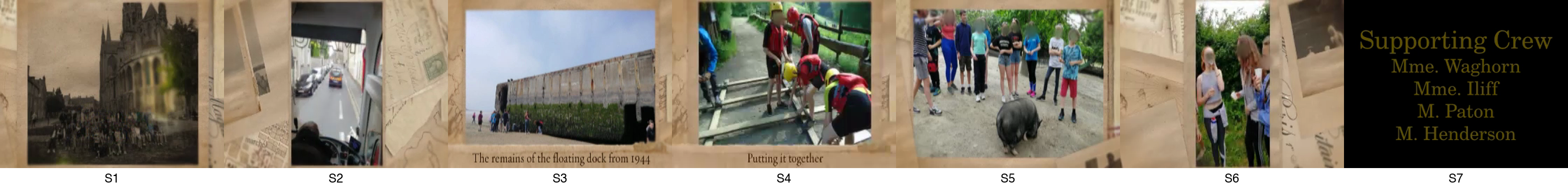} \\
\hline
\begin{minipage}{0.95\linewidth}
\tiny
\textbf{S1} (00:00--00:26): score: 3: Introduction to Mumbai's flooding crisis - Vihar Lake overflowing, torrential rain causing widespread waterlogging, schools and colleges shut in Raigad and Thane districts, local trains running late due to reduced visibility.\\
\textbf{S2} (00:42--01:14): score: 3: Impact assessment showing over 60mm rainfall recorded, NDRF teams deployed, over 60 incidents of tree collapses, wall collapses and short circuits reported. Air India issues travel advisory warning of flight delays. \\
\textbf{S3} (01:36--02:03): score: 2: Weather forecast details - rainfall five times above normal levels (60-68mm vs typical 12-15mm), red alert to reduce to orange category tomorrow, heavy rain expected for next 24-48 hours. \\
\textbf{S4} (03:18--03:50): score: 3: Ground report from Mumbai showing severely impacted areas including Andheri subway (shut due to flooding), Sion, Chembur, and Kurla. Airlines including SpiceJet issuing caution about delays.\\
\textbf{S5} (06:06--06:49): score: 2: School closures across Maharashtra - schools shut in Palghar, Thane, Raigad, and Pune. Pimpri Chinchwad severely impacted with rainwater entering homes.\\
\textbf{S6} (08:00--08:32): score: 2: Second ground report from Parel showing continuous heavy rain for 5-6 hours, BMC officials keeping manholes clean to prevent waterlogging, Andheri subway remains closed."
\end{minipage} & \begin{minipage}{0.95\linewidth}
\tiny
\textbf{S1} (00:09--00:29): score: 3: Opening montage showing students on a ferry deck during the trip, followed by a large group photo in front of a historic cathedral/building, establishing the school trip context.\\
\textbf{S2} (00:29--00:48): score: 2: Police escort segment with text explaining "Had a blue-light escort as roads were blocked!" showing vehicles and French town architecture. \\
\textbf{S3} (00:58--01:18): score: 2: Beach visit segment showing what appears to be D-Day landing beaches in Normandy with historical significance, plus coastal views.\\
\textbf{S4} (01:27--01:57): score: 3: Outdoor adventure activities showing students in safety helmets building rafts ("Putting it together" and "Almost ready to launch" captions), demonstrating team-building exercises.\\
\textbf{S5} (02:16--02:36): score: 2: Farm/animal interaction segment featuring "Kevin the peacock showing off," horses with foal, and students interacting with animals. \\
\textbf{S6} (02:45--03:05): score: 2: Meal time and social activities with captions "Down the hatch" and "the cakes were great too!" showing students enjoying food together.\\
\textbf{S7} (15:07--15:26): score: 3: End credits showing participant names and "Director and Producer: Mme. Trezeux," confirming this is a school-produced trip documentary.
\end{minipage} \\
\noalign{\hrule height 2pt}
\multicolumn{2}{|c|}{\textbf{Gemini 2.5 Pro}} \\
\hline
\includegraphics[width=\linewidth,keepaspectratio]{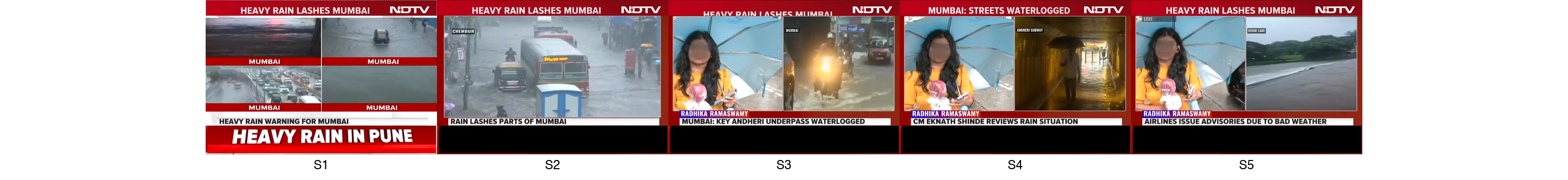} & \includegraphics[width=\linewidth,keepaspectratio]{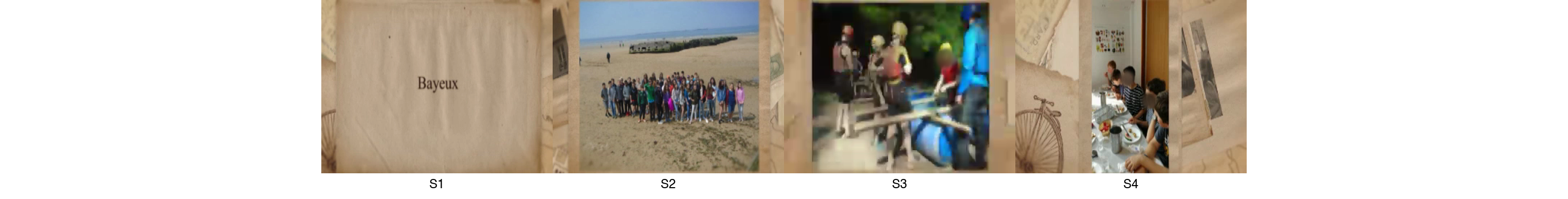} \\
\hline
\begin{minipage}{0.95\linewidth}
\tiny
\textbf{S1} (00:00--00:23): score: 3: An introduction to the news report, stating that torrential rain has caused the Vihar Lake to overflow, shut down the Andheri subway, and led to widespread flooding in Mumbai. \\
\textbf{S2} (02:03--02:29): score: 3: This segment quantifies the severity of the rainfall, explaining that the recorded 60-68mm is more than five times the normal amount for a monsoon day in July, leading to massive waterlogging. \\
\textbf{S3} (03:56--04:12): score: 2: A field reporter details the impact on transportation, highlighting immense traffic on the Western and Eastern Express Highways and advisories from airlines due to the severe weather. \\
\textbf{S4} (05:29--05:45): score: 2: A concise summary of how the heavy rains are affecting the common person, with widespread flooding, traffic congestion, and disruptions to both local train services and air travel. \\
\textbf{S5} (06:23--06:34): score: 2: An update on school closures, specifying that schools in the surrounding areas of Palghar, Thane, Raigad, and Pune have been shut down due to the intense rainfall.
\end{minipage} & \begin{minipage}{0.95\linewidth}
\tiny
\textbf{S1} (00:00--00:24): score: 3: The video opens with a title card, ``Normandy 2018,`` showing a ferry journey, establishing the setting for a JMHS Year 10 school trip. This is followed by a montage of group photos at various locations, including a large cathedral. \\
\textbf{S2} (01:02--01:22): score: 3: Students are shown visiting one of the historic D-Day landing beaches, exploring the remains of a floating dock from 1944 and taking a group photo on the sand. \\
\textbf{S3} (01:31--01:51): score: 2: A montage of students engaging in outdoor adventure activities. They are seen getting kitted out in safety gear and building a raft from barrels. \\
\textbf{S4} (03:27--03:56): score: 2: This segment highlights cultural experiences, showing students filling in their journals and bravely trying local cuisine, including snails.
\end{minipage} \\
\noalign{\hrule height 2pt}
\multicolumn{2}{|c|}{\textbf{Qwen3VL-235B}} \\
\hline
\includegraphics[width=\linewidth,keepaspectratio]{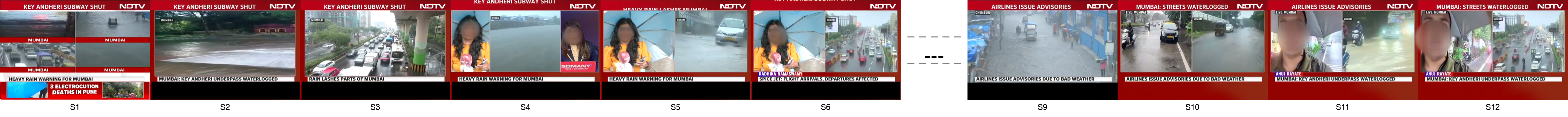} & \includegraphics[width=\linewidth,keepaspectratio]{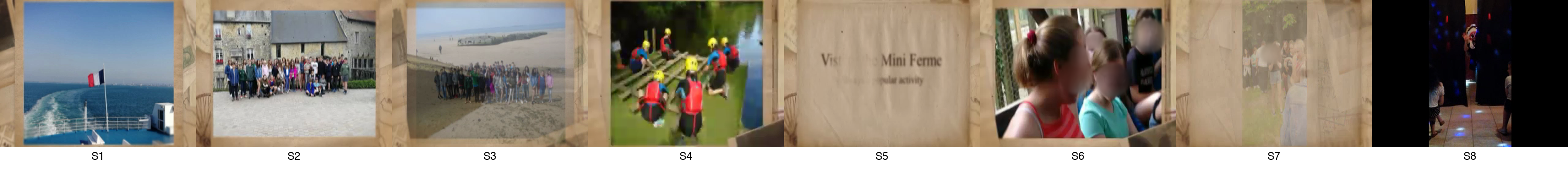} \\
\hline
\begin{minipage}{0.95\linewidth}
\tiny
\textbf{S1} (00:00--00:19): score: 3: Introduction to the severe weather event in Mumbai, highlighting Vihar Lake overflow, widespread waterlogging, and disruption to daily life including subway closures and school shutdowns in key districts.\\
\textbf{S2} (00:35--00:50): score: 3: Quantification of rainfall (over 60mm in Santa Cruz and Colaba), deployment of NDRF teams, and reporting of over 60 incidents including tree and wall collapses, emphasizing the scale of the emergency. \\
\textbf{S3} (01:00--01:14): score: 2: Air travel disruptions with advisories from Air India and SpiceJet, urging travelers to arrive early due to traffic and waterlogging, indicating cascading effects on transportation networks.\\
\textbf{S4} (02:48--03:07): score: 3: Live reporter Radhika Ramaswamy details ground-level impact, identifying severely affected areas like Andheri, Sion, and Chembur, and confirming subway closures and traffic gridlock.\\
\textbf{S5} (03:24--03:37): score: 2: Visual and verbal confirmation of Andheri subway flooding and closure, along with other arterial roads and junctions affected, reinforcing the infrastructure strain.\\
\textbf{S6} (04:23--04:36): score: 2: Coordination of emergency response with NDRF, Army, Navy, and DMC officials working round the clock, and mention of active water pumping efforts in some areas."\\
\textit{---} \\
\textbf{S9} (07:47--08:00): score: 3: On-the-ground reporter Anuj Rayate confirms persistent heavy rain in Perel, noting intermittent intensity and ongoing civic efforts to manage manholes and drainage.\\
\textbf{S10} (08:55--09:07): score: 2: Comparison to previous rainfall events, noting improved waterlogging management except in low-lying Andheri subway, and highlighting civic administration\u2019s proactive measures. \\
\textbf{S11} (09:19--09:36): score: 2: Societal impact including slower pace of life, increased work-from-home, and public advisories to stay indoors, reflecting behavioral changes due to weather conditions.\\
\textbf{S12} (09:47--10:07): score: 2: ASummary of ongoing challenges: traffic slowness, reduced visibility, and potential worsening during evening peak hours, despite improved drainage efforts.
\end{minipage} & \begin{minipage}{0.95\linewidth}
\tiny
\textbf{S1} (00:00--00:17): score: 3: The video opens with a strong, critical monologue in French, targeting political figures and societal norms, setting a rebellious and disillusioned tone. \\
\textbf{S2} (00:41--00:51): score: 2: The speaker expresses personal anxiety and societal pressure, contrasting the image of a \u201cdynamic young guard\u201d with the reality of being a stressed, panicked youth.\\
\textbf{S3} (01:05--01:16): score: 3: A direct critique of media manipulation and societal control, with lines like \u201cLes m\u00e9dias font trop de chantage\u201d and \u201cOn va tous finir en cage,\u201d highlighting themes of oppression and loss of freedom. \\
\textbf{S4} (01:37--01:44): score: 2: Commentary on economic hardship and the relentless grind of work to earn money, using vivid imagery like \u201ccrache, crache et du cash\u201d and \u201cse saigne jusqu'\u00e0 \u00eatre balafr\u00e9.\u201d\\
\textbf{S5} (02:07--02:23): score: 3: The speaker vents frustration at media and public figures, stating they turn on the TV \u201cpour cracher sur des connards que j'd\u00e9teste,\u201d emphasizing anger and disillusionment.\\
\textbf{S6} (02:34--02:41): score: 2: A reiteration of the core message: \u201cJ'n'dis pas c'que j'pense mais j'pense c'que j'dis,\u201d questioning the value of life in a society that feels dishonest and oppressive.\\
\textbf{S7} (03:03--03:15): score: 1: A repetitive, almost chant-like segment acknowledging hypocrisy and envy in society, with the line \u201cC'est vrai qu'on est un petit peu des deux.\u201d\\
\textbf{S8} (10:20--10:28): score: 1: A brief countdown (\u201c4, 3, 2, 1\u201d) appears, possibly signaling a transition or climax, though its context is unclear without further visual or audio cues.
\end{minipage} \\
\hline
\end{tabular}
}
\end{figure*}

%% file: main.bib
@inproceedings{gygli2014creating,
  title     = {Creating Summaries from User Videos},
  author    = {Gygli, Michael and Grabner, Helmut and Riemenschneider, Hayko and Van Gool, Luc},
  booktitle = {Computer Vision--ECCV 2014},
  pages     = {505--520},
  year      = {2014},
  publisher = {Springer International Publishing}
}

@inproceedings{Song2015TVSumSW,
  title={TVSum: Summarizing web videos using titles},
  author={Yale Song and Jordi Vallmitjana and Amanda Stent and Alejandro Jaimes},
  booktitle={Proceedings of the IEEE Conference on Computer Vision and Pattern Recognition (CVPR)},
  pages={5179--5187},
  year={2015}
}

@article{deAvila2011VSUMM,
  title     = {VSUMM: A mechanism designed to produce static video summaries and a novel evaluation method},
  author    = {de Avila, Sandra Eliza Fontes and Lopes, Ana Paula Brand{\~a}o and da Luz Jr, Antonio and de Albuquerque Ara{\'u}jo, Arnaldo},
  journal   = {Pattern Recognition Letters},
  volume    = {32},
  number    = {1},
  pages     = {56--68},
  year      = {2011},
  publisher = {Elsevier},
  doi       = {10.1016/j.patrec.2010.08.004}
}

@article{lin2023videoxum,
  title     = {VideoXum: Cross-modal Visual and Textural Summarization of Videos},
  author    = {Lin, Jingyang and Hua, Hang and Chen, Ming and Li, Yikang and Hsiao, Jenhao and Ho, Chiuman and Luo, Jiebo},
  journal   = {IEEE Transactions on Multimedia},
  volume    = {26},
  pages     = {5548--5560},
  year      = {2024},
  publisher = {IEEE},
  doi       = {10.1109/TMM.2023.3335875}
}

@article{ghauri2020eduvsum, 
   title={Classification of Important Segments in Educational Videos using Multimodal Features},
   author={Ghauri, Junaid Ahmed and Hakimov, Sherzod and Ewerth, Ralph}, 
   Conference={International Workshop on Investigating Learning During Web Search (IWILDS 2020) co-located with CIKM}, 
   year={2020} 
}

@inproceedings{mylonas2025sdvsum,
  title     = {SD-VSum: A Method and Dataset for Script-Driven Video Summarization},
  author    = {Mylonas, Manolis and Apostolidis, Evlampios and Mezaris, Vasileios},
  booktitle = {Proceedings of the 33rd ACM International Conference on Multimedia (ACM MM '25)},
  pages     = {6596--6604},
  year      = {2025},
  publisher = {ACM},
  doi       = {10.1145/3746027.3755821}
}

@inproceedings{liu-etal-2025-talk,
    title = "What Is That Talk About? A Video-to-Text Summarization Dataset for Scientific Presentations",
    author = "Liu, Dongqi  and
      Whitehouse, Chenxi  and
      Yu, Xi  and
      Mahon, Louis  and
      Saxena, Rohit  and
      Zhao, Zheng  and
      Qiu, Yifu  and
      Lapata, Mirella  and
      Demberg, Vera",
    editor = "Che, Wanxiang  and
      Nabende, Joyce  and
      Shutova, Ekaterina  and
      Pilehvar, Mohammad Taher",
    booktitle = "Proceedings of the 63rd Annual Meeting of the Association for Computational Linguistics (Volume 1: Long Papers)",
    month = jul,
    year = "2025",
    address = "Vienna, Austria",
    publisher = "Association for Computational Linguistics",
    url = "https://aclanthology.org/2025.acl-long.310/",
    pages = "6187--6210",
    ISBN = "979-8-89176-251-0"
}

@inproceedings{chen2017video,
  title     = {Video to Text Summary: Joint Video Summarization and Captioning with Recurrent Neural Networks},
  author    = {Chen, Bor-Chun and Chen, Yan-Ying and Chen, Francine},
  booktitle = {Proceedings of the British Machine Vision Conference (BMVC)},
  year      = {2017},
  doi       = {10.5244/C.31.139}
}

@inproceedings{he2023align,
  title     = {Align and Attend: Multimodal Summarization with Dual Contrastive Losses},
  author    = {He, Bo and Wang, Jun and Qiu, Jielin and Bui, Trung and Shrivastava, Abhinav and Wang, Zhaowen},
  booktitle = {Proceedings of the IEEE/CVF Conference on Computer Vision and Pattern Recognition (CVPR)},
  pages     = {14867--14878},
  year      = {2023},
  doi       = {10.1109/CVPR52729.2023.01428}
}

@inproceedings{huang2020query,
  title     = {Query-controllable Video Summarization},
  author    = {Huang, Jia-Hong and Worring, Marcel},
  booktitle = {Proceedings of the 2020 International Conference on Multimedia Retrieval (ICMR '20)},
  pages     = {242--250},
  year      = {2020},
  publisher = {Association for Computing Machinery},
  address   = {New York, NY, USA},
  doi       = {10.1145/3372278.3390695}
}

@inproceedings{narasimhan2022tl,
  title={TL; DW? Summarizing Instructional Videos with Task Relevance and Cross-Modal Saliency},
  author={Narasimhan, Medhini and Nagrani, Arsha and Sun, Chen and Rubinstein, Michael and Darrell, Trevor and Rohrbach, Anna and Schmid, Cordelia},
  booktitle={European Conference on Computer Vision},
  pages={540--557},
  year={2022},
  organization={Springer}
}

@inproceedings{Narasimhan2021CLIPIt,
  title     = {CLIP-It! Language-Guided Video Summarization},
  author    = {Narasimhan, Medhini and Rohrbach, Anna and Darrell, Trevor},
  booktitle = {Advances in Neural Information Processing Systems (NeurIPS)},
  volume    = {34},
  pages     = {13988--14000},
  year      = {2021}
}

@inproceedings{Sharghi2016QueryFocused,
  title     = {Query-Focused Extractive Video Summarization},
  author    = {Sharghi, Aidean and Gong, Boqing and Shah, Mubarak},
  booktitle = {European Conference on Computer Vision (ECCV)},
  pages     = {3--19},
  year      = {2016},
  publisher = {Springer}
}

@inproceedings{wei2018video,
  title     = {Video Summarization via Semantic Attended Networks},
  author    = {Wei, Huawei and Ni, Bingbing and Yan, Yichao and Yu, Huanyu and Yang, Xiaokang and Yao, Chen},
  booktitle = {Proceedings of the AAAI Conference on Artificial Intelligence},
  volume    = {32},
  number    = {1},
  pages     = {216--223},
  year      = {2018},
  url       = {https://doi.org/10.1609/aaai.v32i1.11297}
}

@inproceedings{lee2025video,
  title     = {Video Summarization with Large Language Models},
  author    = {Lee, Min Jung and Gong, Dayoung and Cho, Minsu},
  booktitle = {Proceedings of the IEEE/CVF Conference on Computer Vision and Pattern Recognition (CVPR)},
  year      = {2025},
  url       = {https://arxiv.org/abs/2504.11199}
}

@inproceedings{hua2025v2xum,
  title     = {V2Xum-LLM: Cross-Modal Video Summarization with Temporal Prompt Instruction Tuning},
  author    = {Hua, Hang and Tang, Yunlong and Xu, Chenliang and Luo, Jiebo},
  booktitle = {Proceedings of the AAAI Conference on Artificial Intelligence},
  volume    = {39},
  number    = {4},
  pages     = {3599--3607},
  year      = {2025},
  doi       = {10.1609/aaai.v39i4.32374},
  url       = {https://ojs.aaai.org/index.php/AAAI/article/view/32374}
}

@inproceedings{otani2019rethinking,
  title     = {Rethinking the Evaluation of Video Summaries},
  author    = {Otani, Mayu and Nakashima, Yuta and Rahtu, Esa and Heikkil{\"a}, Janne},
  booktitle = {Proceedings of the IEEE/CVF Conference on Computer Vision and Pattern Recognition (CVPR)},
  pages     = {7596--7604},
  year      = {2019},
  doi       = {10.1109/CVPR.2019.00777}
}

@inproceedings{krishna2017dense,
  title={Dense-Captioning Events in Videos},
  author={Krishna, Ranjay and Hata, Kenji and Ren, Frederic and Fei-Fei, Li and Niebles, Juan Carlos},
  booktitle={Proceedings of the IEEE International Conference on Computer Vision (ICCV)},
  pages={706--715},
  year={2017},
  doi={10.1109/ICCV.2017.83}
}

@article{gemini25report2025,
  title={Gemini 2.5: Pushing the Frontier with Advanced Reasoning, Multimodality, Long Context, and Next Generation Agentic Capabilities},
  author={Comanici, Gheorghe and Bieber, Eric and Schaekermann, Mike and Pasupat, Ice and Sachdeva, Noveen and Dhillon, Inderjit and Blistein, Marcel and Ram, Ori and others},
  journal={arXiv preprint arXiv:2507.06261},
  year={2025},
  url={https://arxiv.org/abs/2507.06261}
}

@article{qwen3vlreport2025,
  title={Qwen3-VL Technical Report},
  author={Bai, Shuai and Cai, Yuxuan and others},
  journal={arXiv preprint arXiv:2511.21631},
  year={2025},
  url={https://arxiv.org/abs/2511.21631}
}

@article{Apostolidis2021VideoSU,
  author    = {Apostolidis, Evampelos and Belaid, Eleni and Mezaris, Vasileios and Patras, Ioannis},
  title     = {Video Summarization Using Deep Learning: A Survey},
  journal   = {IEEE Transactions on Circuits and Systems for Video Technology},
  volume    = {31},
  number    = {7},
  pages     = {2873--2891},
  year      = {2021},
  doi       = {10.1109/TCSVT.2020.3032165}
}

@misc{OpenAI2025GPT5Science,
  author       = {OpenAI},
  title        = {Early experiments in accelerating science with GPT-5},
  year         = {2025},
  eprint       = {2511.16072},
  archivePrefix = {arXiv},
  primaryClass  = {cs.CL},
  url          = {https://arxiv.org/abs/2511.16072},
  doi          = {10.48550/arXiv.2511.16072}
}

@misc{anthropic2025opus45,
  title  = {Introducing Claude Opus 4.5},
  author = {{Anthropic}},
  year   = {2025},
  url    = {https://www.anthropic.com/news/claude-opus-4-5},
  note   = {Accessed: 2026-02-14}
}
